\documentclass[acmsmall,nonacm,screen,11pt]{acmart}
\usepackage[layout=paper]{appliedpaper}
\usepackage{booktabs}
\usepackage{tabularx}
\usepackage{listings}
\usepackage{wrapfig}
\usepackage{ulem}
\usepackage{xspace}
\newsavebox{\transitionQualBox}

\title[STRIKE: Learning Visual State Transitions for Physical World Modeling]{STRIKE: Learning Visual State Transitions for Physical World Modeling}
\DeclareRobustCommand{\ours}{STRIKE\xspace}

\author{Wenbin Teng}
\affiliation{\institution{Applied Intuition}}
\affiliation{\institution{University of Southern California}}
\author{Tianshuo Xu}
\affiliation{\institution{Applied Intuition}}
\author{Depu Meng}
\affiliation{\institution{Applied Intuition}}
\author{Yuelei Li}
\affiliation{\institution{Applied Intuition}}
\author{Quentin Herau}
\affiliation{\institution{Applied Intuition}}
\author{Yihan Hu}
\affiliation{\institution{Applied Intuition}}
\author{Yajie Zhao}
\affiliation{\institution{University of Southern California}}
\author{Wei Zhan}
\affiliation{\institution{Applied Intuition}}
\affiliation{\institution{University of California, Berkeley}}
\email{wei.zhan@applied.co}

\compactauthors{%
  {\raggedright
   {\rmfamily\bfseries
    \appliedname{Wenbin Teng\textsuperscript{1,2,\ast}}, \;
    \appliedname{Tianshuo Xu\textsuperscript{1}}, \;
    \appliedname{Depu Meng\textsuperscript{1,\S}}, \;
    \appliedname{Yuelei Li\textsuperscript{1}},\\[0.25em]
    \appliedname{Quentin Herau\textsuperscript{1}}, \;
    \appliedname{Yihan Hu\textsuperscript{1}}, \;
    \appliedname{Yajie Zhao\textsuperscript{2}}, \;
    \appliedname{Wei Zhan\textsuperscript{1,3,\dag}}\par}
   \smallskip
   {\small\rmfamily\color{AppliedGray}%
    \textsuperscript{1}Applied Intuition \quad
    \textsuperscript{2}University of Southern California \quad
    \textsuperscript{3}University of California, Berkeley\par}%
   \smallskip
   {\footnotesize\rmfamily\color{AppliedGray}%
   \textsuperscript{\ast}Work done during an internship at Applied Intuition.\quad
    \textsuperscript{\S}Project Lead.\quad
    \textsuperscript{\dag}Corresponding author:
    \href{mailto:wei.zhan@applied.co}{wei.zhan@applied.co}\par}
    {\footnotesize\rmfamily\color{AppliedGray}%
    Project Page:
    \href{https://strike-applied.github.io/}{https://strike-applied.github.io/}\par}
   \par}
}

\setpaperteaser{%
  \centering
  \paperteasergraphic{\includegraphics[width=0.97\linewidth]{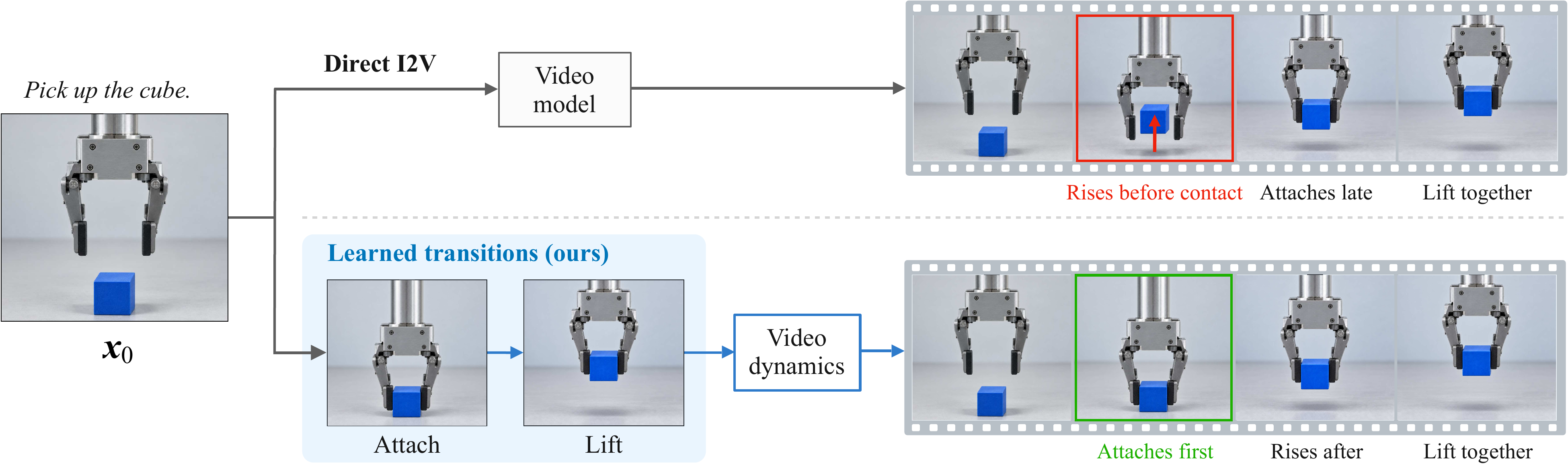}}
  \caption{\textbf{Overview.} Given an initial image and task, \ours uses a timed event plan to recursively predict future visual states with a transition model learned from event-aligned state pairs. The predicted states and their temporal positions are applied as conditions of dense video generation.}
  \label{fig:teaser_transitions}
}

\begin{document}
% !TEX root = main.tex

\begin{abstract}
    % Physically meaningful video prediction requires realizing correct interaction outcomes and generating temporally coherent motion around them. We investigate interaction-state realization and conditional trajectory realization as two coupled prediction demands. By comparing video generation with and without interaction-state conditions, we find that accurate states improve downstream interaction consistency. Motivated by this analysis, we learn a conditional transition model that predicts a future visual state from the current observation, a transition specification, elapsed time, and optional embodiment conditions. The model is applied successively to construct a timed sequence of visual states, which conditions a video model to generate dense motion. Experiments on VideoPhy, PhyGenBench, Physics-IQ Verified, and RoboTwin 2.0 show that learning visual state transitions improves interaction-state accuracy, supporting more reliable physical world modeling.
    Physical world modeling requires predicting how interactions change a scene, not merely generating coherent motion. We propose \ours, a framework that separates visual state transition learning from dense video generation. We construct event-aligned supervision by extracting observed states from training videos and pairing them with transition descriptions and temporal offsets. An image-based transition model learns to predict the next scene configuration from the current image, a local transition specification, and elapsed time. At inference, a pretrained vision-language planner predicts time transition specifications, and recursive application of the learned transition model produces a sequence of future visual states. A separately trained dynamic model then generates the complete rollout conditioned on these states and their temporal locations. Experiments on Physics-IQ Verified, PhyGenBench, Pisa-Experiments, and RoboTwin2.0 show improvements of \ours over the corresponding video-backbone baselines in benchmark measures of physical consistency and manipulation-video fidelity. These results support learned visual state transitions as an effective intermediate representation for physical world modeling.
    % Physical world modeling requires predicting how interactions change a scene, not merely generating coherent motion. We propose \ours, a framework that
    % explicitly learns visual state transitions and uses the predicted states to condition dense video generation. At its core is an image-based transition model trained on paired visual states to predict the next scene configuration from the current image, a transition specification, and elapsed time. A pretrained vision-language model supplies timed transition specifications, and recursive application of the learned transition model produces a sequence of future visual states. A conditional video diffusion model then generates the full rollout from these states and their temporal locations. This separates supervised state-transition learning from conditional dynamics modeling. Experiments on Physics-IQ Verified, PhyGenBench, Pisa-Experiments, and RoboTwin2.0 show improvements over the corresponding video-backbone baselines in benchmark measures of physical realism and manipulation-video fidelity. These results support learned visual state transitions as an effective intermediate representation for physical world modeling.
\end{abstract}

\makefrontmatter
\insertpaperteaser

% !TEX root = main.tex

\section{Introduction}

A central challenge in physical world modeling is predicting the consequences of interactions among objects and agents. For example, a successful grasp establishes attachment between an object and a gripper, while lift preserves that attachment and changes the object's support conditions (Figure~\ref{fig:teaser_transitions}). We use visual states to denote configurations that express such interaction outcomes~\citep{zeng2024learning}. These states capture changes in object configurations and relations that constrain subsequent physical evolution. Preserving object appearance alone is therefore insufficient: the generated scene must also establish the relation of transition that make an interaction physically meaningful.

We consider a formulation that treats transitions between visual states as an explicit learning target. A state-transition model predicts the state reached after an interaction, given the current observation and an interaction specification. A conditional dynamics model then models how the scene evolves between the predicted states. For open-domain generation, a pretrained vision-language event planner provides the local transition specifications and their temporal locations. This formulation allows us to investigate whether explicitly learning state transitions improves the prediction of interaction outcomes and their subsequent physical consequences.

% This perspective motivates treating transitions between visual states as an explicit supervised learning target. Event-aligned pairs of observed states provide supervision for how a specified physical change is visually realized. Predicted transition states can then serve as intermediate conditions for dense video generation, separating the prediction of meaningful scene configurations from the synthesis of continuous temporal evolution. We investigate whether this explicit transition-learning formulation improves physical video generation across different phenomena and remains effective with different video backbones.

Learning state transitions is a general approach to predictive modeling for planning and control. Latent dynamics models support planning and behavior learning~\citep{hafner2019planet,hafner2020dreamer}, visual prediction supports robot manipulation~\citep{ebert2018visualforesight}, and language-conditioned video models represent task execution through visual plans~\citep{du2023unipi}. Within this broader setting, we study a transition interface defined by the current image, a local description of a physical change, and elapsed time. Our focus is on learning this mapping from event-aligned state pairs and composing its predictions to guide dense visual evolution.

% The paper layout renders this figure on its opening page.
In this work, we propose \ours, a novel framework that explicitly learns visual state transitions and uses the predicted states to condition dense video generation. We first instantiate the state-transition model as a conditional image model that predicts a future visual state from the current observation, an interaction specification, and the elapsed time. To construct its supervision, a VLM-assisted curation pipeline identifies events in training videos, selects informative visual states, and annotates the transitions between them. After structural and quality checks, adjacent selected states are paired with transition descriptions and elapsed times, yielding supervision grounded in observed changes rather than generic video-level captions. The selected images are extracted from training videos rather than synthesized by the annotator. We train the image model on these pairs, directly supervising how a specified interaction changes the scene's object configurations and relations. During inference, the event planner instead predicts future transition conditions from the initial image and task context, without observing future reference frames. Each predicted state becomes the input for the next specified interaction, allowing the model to construct a sequence of successive visual states. These predictions provide explicit state and timing conditions for dense video generation. Second, the conditional dynamics model is implemented as a video diffusion model trained to generate the motion and intermediate configurations connecting these states. Together, the two components model physical evolution through explicit interaction outcomes and the dynamics between them. 

Our contributions are as follows:
  \begin{itemize}
      \item We formulate visual state-transition prediction as an explicit event- and time-conditioned learning task, and
      construct event-aligned supervision from observed video states through a VLM-assisted annotation pipeline.

      \item We develop a compositional generation framework that combines timed event planning with recursive prediction of
      transition states. These states and their temporal locations provide soft conditions for separately trained video
      dynamics models, instantiated with CogVideoX and Wan.

      \item We evaluate the framework on Physics-IQ Verified, PhyGenBench, PisaBench, and RoboTwin2.0, demonstrating gains in
      physical-consistency and manipulation-video metrics over the corresponding backbone baselines.
  \end{itemize}

\section{Related Work}
\subsection{Physics-Aware Generative World Models.}
Large-scale video generators, including CogVideoX~\citep{yang2025cogvideox}, Wan~\citep{wan2025wan}, and Cosmos~\citep{ali2025world,agarwal2026cosmos}, provide backbones for visual world modeling. IRASim~\citep{zhu2025irasim} aligns action trajectories with video frames, while Unified World Models~\citep{zhu2025unified} jointly models video and action diffusion for dynamics and policy learning. Complementary approaches target physical fidelity through physics-aware conditioning~\citep{wang2025wisa}, simulation-based post-training~\citep{li2025pisa}, relational alignment~\citep{zhang2025videorepa}, latent motion priors~\citep{jiang2026lamo}, and staged motion control~\citep{xu2026motion}. These approaches primarily address video-level generation or control; we focus on explicitly predicting the outcomes of individual interactions.

% !TEX root = main.tex

\begin{figure*}[t]
  \centering
  \includegraphics[width=0.99\textwidth]{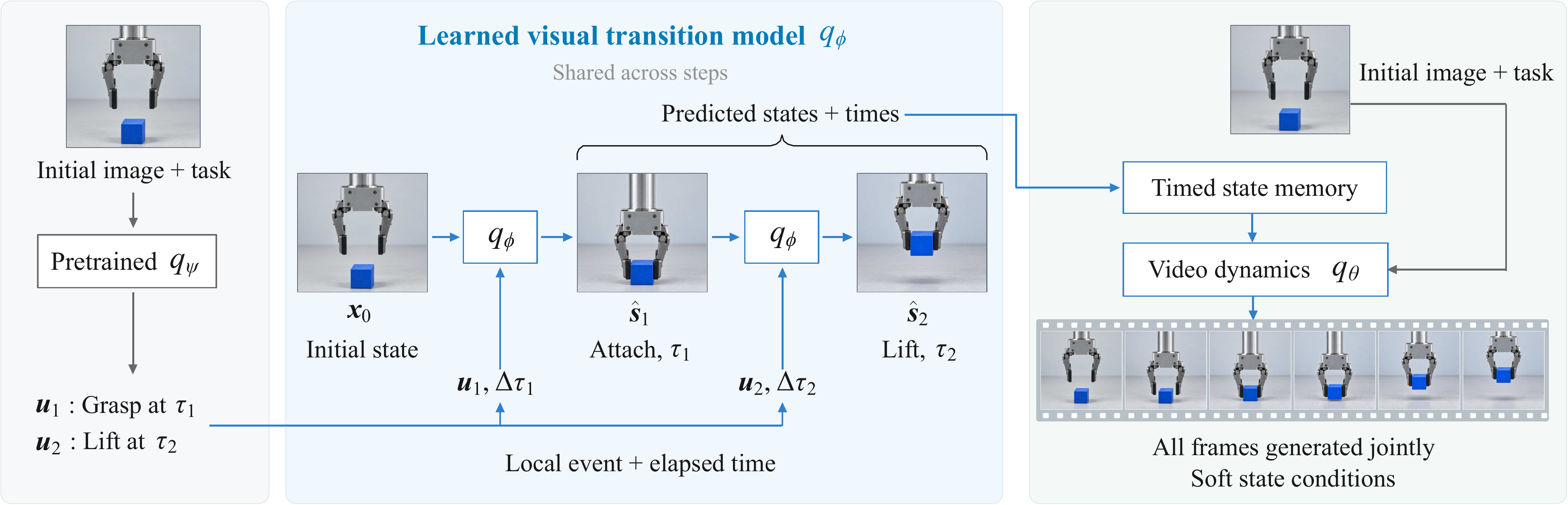}
  \caption{\textbf{The overall framework of \ours.} (a) A pretrained vision-language event planner receives the initial image $\symbfit{x}_0$ and task $\symbfit{c}$, and proposes local events $\symbfit{u}_k$ and outcome times $\tau_k$. (b) A shared visual transition model $q_\phi$ is applied successively using the current state, event, and elapsed time $\Delta\tau_k=\tau_k-\tau_{k-1}$. Observed event-aligned state pairs supervise its training. (c) A separately trained dynamics model $q_\theta$ generates the dense rollout from the initial context and soft conditions supplied by the predicted states and their temporal locations.}
  \label{fig:arch}
\end{figure*}

\subsection{Keyframe- and Event-Guided Generation.}
KeyWorld~\citep{li2025keyworld} generates task-relevant keyframes from an initial observation and a global task description before reconstructing dense video. SKIP~\citep{he2026skip} learns event-preserving sparse rollouts with interpolation. DCARL~\citep{ouyang2026dcarl} trains a keyframe generator to establish consistent structural anchors for long trajectory generation. RoboEnvision~\citep{yang2025roboenvision} decomposes manipulation tasks into atomic instructions, generates aligned boundary keyframes, and fills the intervening segments conditioned on those instructions. CausalMotion~\citep{zhuang2026causalmotion} uses VLM-planned keyframes and object trajectories to guide a pretrained video generator without additional training, while CoECT~\citep{wang2026coect} constructs event keyframes through event-chain reasoning and iterative editing. We instead directly supervise a shared transition model on event-aligned state pairs, conditioned on the current visual state, one interaction specification, and elapsed time. 

\subsection{Visual Subgoal and State-Transition Prediction.}
Visual state prediction also supports manipulation representation learning and policy guidance. MPI~\citep{zeng2024learning} pretrains representations through interaction-frame prediction and object localization. SuSIE~\citep{black2024susie} uses an image-editing diffusion model to propose language-conditioned subgoals for a low-level policy, while TaKSIE~\citep{kang2025tasksie} incorporates task progress for adaptive subgoal generation. Rather than learning representations or subgoals, we supervise interaction-conditioned transitions across physical phenomena and use the timed visual outcomes to condition dense video dynamics.

% !TEX root = main.tex

\section{Preliminaries}
\label{sec:preliminaries}

\subsection{Visual States and Transition Conditions}

We consider generating a visual rollout $\symbfit{X}=(\symbfit{x}_0,\symbfit{x}_1,\ldots,\symbfit{x}_{F-1})$ from an initial visual condition $\symbfit{x}_0$ and task context $\symbfit{c}$, such as a scene description or a manipulation instruction. For image-conditioned generation, $\symbfit{x}_0$ is an observed frame; text-only generation additionally requires synthesizing an initial image before this rollout stage.

We use a \emph{visual state} to denote an image depicting a scene configuration, including observable object properties and relations such as contact, support, and attachment. A state can depict either an interaction boundary or its persistent consequence, such as first contact with a surface or an object resting after landing. Unlike a complete simulator state, a visual state does not explicitly encode quantities such as velocity or occluded contacts.

We distinguish a visual state $\symbfit{s}$ from a \emph{transition specification} $\symbfit{u}$, which describes a local change and its anticipated visible outcome. For example, a release transition may specify that an object becomes detached from a gripper; the corresponding state image depicts how that outcome is realized in the particular scene. The global context $\symbfit{c}$ describes the overall task or process, whereas $\symbfit{u}$ specifies an individual step.

Let $\symbfit{S}=(\symbfit{s}_1,\ldots,\symbfit{s}_K)$ denote $K$ future visual states at ordered video timestamps $\tau_1,\ldots,\tau_K$. We set
$\symbfit{s}_0=\symbfit{x}_0$ and $\tau_0=0$, and define:
\begin{equation}
    0=\tau_0<\tau_1<\cdots<\tau_K,
    \qquad
    \Delta\tau_k=\tau_k-\tau_{k-1}.
\end{equation}
Thus, $\tau_k$ locates a state within the rollout, while $\Delta\tau_k$ specifies the interval between successive states.
The initial observation is not counted among the $K$ future states.

% !TEX root = main.tex

\begin{figure}[!t]
    \centering
    \includegraphics[width=0.99\linewidth]{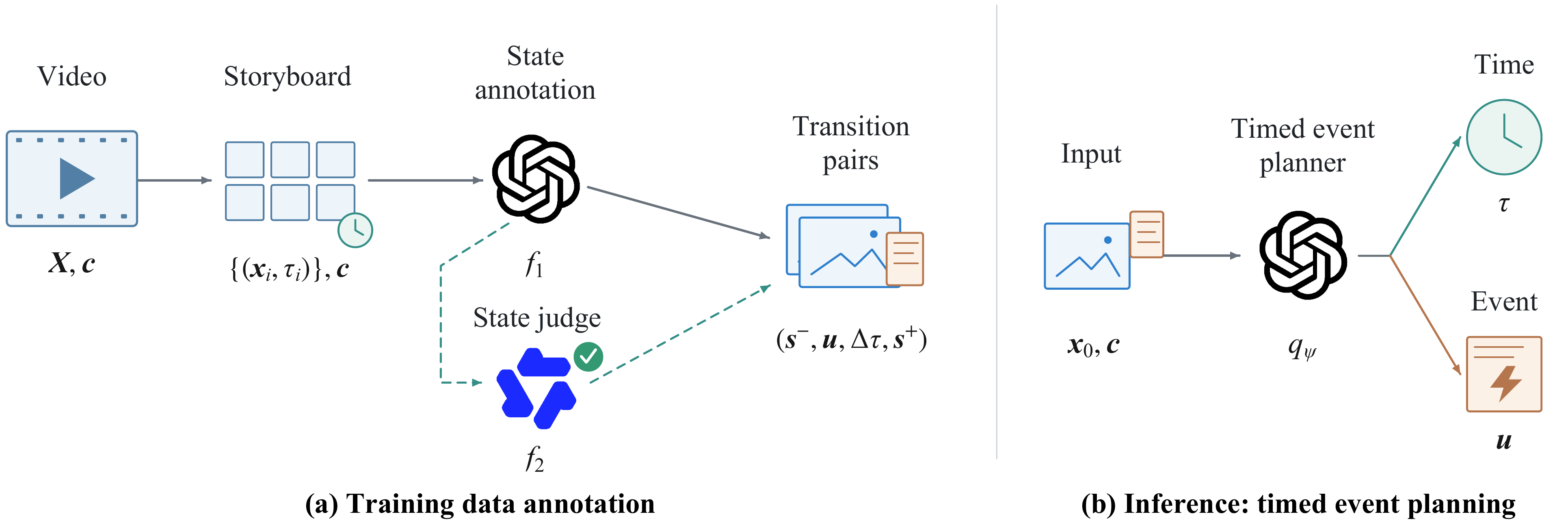}
    \caption{\textbf{Data annotation and timed event planning.} (a) A training video $\symbfit{X}$ and caption $\symbfit{c}$ are converted into a time-labeled storyboard. GPT-5.5~\citep{openai2026gpt55} ($f_1$) selects visual states and annotates their transitions. A Qwen3-VL state judge ($f_2$) checks image--description agreement, with the dashed branch indicating source-specific filtering. Selected video frames and their annotations form transition pairs $(\symbfit{s}^{-},\symbfit{u},\Delta\tau,\symbfit{s}^{+})$. (b) During inference, the event planner $q_{\psi}$ receives an initial image $\symbfit{x}_0$ and task context $\symbfit{c}$, and jointly predicts transition specifications $\symbfit{u}$ and their target times $\tau$, without observing future reference frames.}
    \label{fig:annotation_planning_pipeline}
\end{figure}

\subsection{Conditional Latent Generation}
Conditional image and video generators operate on latent representations of their prediction targets. Let $\symbfit{z}=\mathcal{E}(\symbfit{X})$ denote the latent encoding of an image or video target $\symbfit{X}$, and let $\symbfit{C}$ collect the available conditioning information. A general denoising formulation constructs:
\begin{equation}
    \symbfit{z}_{\lambda}
    =
    a_{\lambda}\symbfit{z}+b_{\lambda}\symbfit\epsilon,
    \qquad
    \symbfit\epsilon\sim\mathcal{\symbfit{N}}(0,\symbfit{I}),
\end{equation}
and optimizes
\begin{equation}
    \mathcal{L}_{\mathrm{gen}}(\eta)
    =
    \mathbb{E}_{\symbfit{z},\symbfit{C},\lambda,\symbfit\epsilon}
    \left[
        w_{\lambda}
        \left\|
            D_{\eta}(\symbfit{z}_{\lambda},\lambda;\symbfit{C})
            -
            \symbfit{z}_{\lambda}^\text{tgt}
        \right\|_2^2
    \right],
\end{equation}
where $a_{\lambda}$ and $b_{\lambda}$ define the corruption process, $D_{\eta}$ is the conditional predictor, and $\symbfit{z}_{\lambda}^\text{tgt}$
denotes the prediction target, which parameterizes the velocity for CogVideoX~\cite{yang2025cogvideox} and Wan~\cite{wan2025wan22}. $w_{\lambda}$ is the backbone-specific noise schedule weighting. The denoising coordinate $\lambda$ is distinct from both the video timestamp $\tau_k$ and the transition interval $\Delta\tau_k$.

Our method retains these generative objectives and introduces structure through the prediction targets and conditioning interface: a transition model predicts sparse visual states under explicit transition and temporal conditions, a video model generates the dense rollout conditioned on those states.

% !TEX root = main.tex

\section{Method}
\label{sec:method}

As illustrated in Figure~\ref{fig:arch}, our framework contains two learned generative components: a visual state-transition model and a state-conditioned video dynamics model. A pretrained vision-language model~\citep{openai2026gpt55} supplies the semantic and temporal conditions that organize their predictions.

\subsection{Data Annotation and Timed Event Planning}
\label{sec:annotation_planning}

We use a common transition interface in two settings: data annotation of training videos and prospective event planning for generation (Figure~\ref{fig:annotation_planning_pipeline}). Annotation describes observed changes to construct supervision; planning predicts the changes to realize from the available initial condition and task context.

\noindent\textbf{Data annotation.}
For each training video $\symbfit{V}$ and its caption $\symbfit{c}$, we sample candidate frames and construct a storyboard labeled with frame indices and timestamps. A first VLM pass identifies the main entities, visible events, and their chronological structure. A second pass selects informative state frames and describes the transitions between adjacent selected states. Selection favors clearly established relations and distinct outcomes, such as contact, deformation, or a settled configuration, rather than generic beginning, middle or end coverage.

We validate candidate identities and temporal order, resolving the selected frames and their timestamps against the candidate table. Quality screening considers event visibility, caption alignment, state coverage, and frame redundancy. Selected state images are extracted from the training videos, not synthesized by the VLM. Adjacent states produce the transition-pair corpus
\begin{equation}
    \mathcal{D}_{\mathrm{trans}}
    = \{(\symbfit{s}^{-},\symbfit{u},\Delta\tau,\symbfit{s}^{+})\},
    \label{eq:annotated_transition_data}
\end{equation}
where $\symbfit{u}$ describes the observed change and $\Delta\tau$ is derived from the selected frame indices on the video timeline. Pairs retain a common video-level grouping identifier for train/validation splitting. This corpus provides the event-aligned supervision in Section~\ref{sec:visual_transitions}. The annotated video/state format also supports the timed-memory supervision in Section~\ref{sec:conditional_dynamics}.

\noindent\textbf{Timed event planning.}
For open-domain inference, a pretrained VLM instead receives only the initial visual condition $\symbfit{x}_0$ and task context $\symbfit{c}$ and predicts
\begin{equation}
    \symbfit{\hat{P}} \sim q_{\psi}(\symbfit{P}\mid \symbfit{x}_0,\symbfit{c}),
    \qquad
    \symbfit{P}=\{(\symbfit{u}_k,\tau_k)\}_{k=1}^{K}.
    \label{eq:method_plan}
\end{equation}
Each $\symbfit{u}_k$ specifies a local change and its anticipated visible outcome, while $\tau_k$ locates that outcome within the rollout. Unlike offline annotation, this stage has no access to future reference frames. We convert the plan into the transition model's instruction format, validate temporal order and horizon bounds, and compute $\Delta\tau_k=\tau_k-\tau_{k-1}$. The descriptions and relative intervals condition successive state predictions; the absolute timestamps position those predictions within the video model. The VLM components are not optimized jointly with either generator. 

\subsection{Learning and Composing Visual State Transitions}
\label{sec:visual_transitions}
We instantiate the transition model $q_{\phi}$ with Qwen-Image-Edit~\citep{wu2025qwen} and fine-tune it on examples $(\symbfit{s}^{-},\symbfit{u},\Delta\tau,\symbfit{s}^{+})$. The source image $\symbfit{s}^{-}$ provides the current scene configuration, $u$ specifies the local change, and $\Delta\tau$ supplies the prediction interval. The target $\symbfit{s}^{+}$ directly supervises the visual configuration resulting from that change.

\noindent\textbf{Relative-time conditioning.}
We encode elapsed time through the temporal coordinates of the image tokens' rotary positional embeddings (RoPE) \citep{su2023roformerenhancedtransformerrotary}, retaining their spatial coordinates. Source and target image tokens receive temporal positions $0$ and $d$, respectively. If a plan specifies frame indices $i_{k-1}$ and $i_k$ on a grid with frame rate $f_{\mathrm{plan}}$, we convert their difference to the canonical training-frame units:
\begin{equation}
    d_k = \operatorname{round}\!\left(
        (i_k-i_{k-1})
        \frac{f_{\mathrm{train}}}{f_{\mathrm{plan}}}
    \right).
    \label{eq:method_relative_time}
\end{equation}
Here, $d_k$ maintains similar representation as $\Delta\tau_k$, not the denoising coordinate $\lambda$. This temporal-position modification introduces no additional trainable parameters.

\noindent\textbf{Transition supervision.}
Let $\symbfit{z}^{+}=\mathcal{E}_I(\symbfit{s}^{+})$ be the target image latent. For Gaussian noise $\symbfit\epsilon$ and noise level $\lambda\in[0,1]$, we construct noisy target latent:
\begin{equation}
    \symbfit{z}^{+}_{\lambda}
    = (1-\lambda)\symbfit{z}^{+}+\lambda\symbfit\epsilon,
    \label{eq:method_transition_noising}
\end{equation}
and the transition model $q_\phi$ is trained to optimize the flow-matching objective
\begin{equation}
    \mathcal{L}_{\mathrm{trans}}(\phi)
    = \mathbb{E}\Bigg[
        \Bigg\|
            \symbfit{v}_{\phi}(\symbfit{z}^{+}_{\lambda},\lambda;\symbfit{s}^{-},\symbfit{s}_0,\symbfit{u},d)
            -(\symbfit\epsilon-\symbfit{z}^{+})
        \Bigg\|_2^2
    \Bigg].
    \label{eq:method_transition_loss}
\end{equation}
The source image remains uncorrupted, and the loss is applied only to target outputs. We also introduce a first-frame ($\symbfit{s}_0$) sink that keeps the latent of the first frame as a persistent visual reference. This helps preserve object identity, scene geometry and appearance over autoregressive transition state generation. We fine-tune the transformer while keeping the pretrained image and multi-modal conditioning encoders frozen.

\noindent\textbf{Sequential composition.} We reuse the learned transition model to materialize the plan:
\begin{equation}
    \hat{\symbfit{s}}_0=\symbfit{x}_0,
    \qquad
    \hat{\symbfit{s}}_k \sim q_{\phi}
    \Bigg(\symbfit{s}_k \Bigg| \hat{\symbfit{s}}_{k-1},\hat{\symbfit{s}}_0,\symbfit{u}_k,\Delta\tau_k\Bigg),
    \quad k=1,\ldots,K.
    \label{eq:method_state_rollout}
\end{equation}
The temporal interval is represented internally by $d_k$ from Eqn.~(\ref{eq:method_relative_time}). Each prediction becomes the visual input to the next transition, with parameters shared across steps. Training uses dataset source images, whereas inference uses the preceding prediction. Although state-generation errors can accumulate during composition, we find adding the first frame as a sink input would effectively alleviate the exposure bias.

\subsection{Video Dynamics Conditioned on Timed Visual States}
\label{sec:conditional_dynamics}
The dynamics model generates the complete video from the predicted states and their temporal locations:
\begin{equation}
    \hat{\symbfit{X}} \sim q_{\theta}\!\Big(
        \symbfit{X}\Big| \symbfit{x}_0,\symbfit{c},
        \{(\hat{\symbfit{s}}_k,\tau_k)\}_{k=1}^{K}
    \Big).
    \label{eq:method_video_rollout}
\end{equation}
We instantiate this model with CogVideoX-5B-I2V~\citep{yang2025cogvideox} and Wan2.2-5B-TI2V~\citep{wan2025wan22}, extending their conditioning interfaces to accept multiple state images. Each predicted state is encoded with the video backbone's VAE and projected into spatial memory tokens. The timestamp $\tau_k$ is mapped to the backbone's temporal grid,
so each memory token carries both spatial and video-time position. Unlike the transition model's relative interval, this position locates the state within the full rollout.

Schematically, video attention accesses the concatenated sequence
\begin{equation}
    \symbfit{H}_{\mathrm{attn}}
    = [\symbfit{M}_1;\ldots;\symbfit{M}_K;\symbfit{H}_{\mathrm{video}}],
    \label{eq:method_memory}
\end{equation}
where $\symbfit{M}_k$ contains the tokens for state $\hat{\symbfit{s}}_k$ and $\symbfit{H}_{\mathrm{video}}$ contains the noisy video tokens. Backbone-specific implementations preserve the original text and initial-image conditioning interfaces. CogVideoX processes a joint
memory/video token stream, while Wan inserts fixed state memories into each block's self-attention and retains separate text
cross-attention. The denoising prediction is read out only from the video-token outputs. 

We retain the backbones' native prediction objectives: scheduler-based velocity prediction for CogVideoX and flow matching for Wan. State images act as soft conditions rather than hard constraints on the corresponding output frames. The model generates the full video window jointly, allowing motion between two states to depend on the wider sequence of predicted outcomes, rather than synthesizing each interval independently.

% !TEX root = main.tex

\section{Experiments}
% !TEX root = main.tex

\begin{table}[t]
  \caption{\textbf{Physics-IQ Verified.} Comparison of model size, generation cost (EFLOPs), and physical consistency across baselines. \textbf{Bold} and \underline{underlined} denote best and second-best scores.}
  \label{tab:physicsiq-verified}
  \begin{center}
  % \resizebox{\linewidth}{!}{%
  \begin{tabular}{lccccccc}
  \toprule
  Model & \# params & EFLOPs & Score $\uparrow$ & S-IoU $\uparrow$ & ST-IoU $\uparrow$ & WS-IoU $\uparrow$ \\
  \midrule
  Wan2.2~\citep{wan2025wan22}         & 14B & 49.100 & 34.2          & \textbf{50.3} & 25.2          & 29.9 \\
  Cosmos-Predict-2.5~\citep{ali2025world} & 2B & 2.560 &32.2          & 45.5          & 27.2          & 30.1 \\
  CogVideoX~\citep{yang2025cogvideox}  & 5B & 6.640 & 30.5          & 40.4          & 30.3          & 24.1 \\
  Cosmos3-Nano~\cite{agarwal2026cosmos} & 16B & 11.802 & 27.6          & 39.7          & 20.2          & 22.1 \\
  CoECT~\citep{wang2026coect}           & 20B+5B & 29.600 & 19.9          & 27.6          & 23.2          & 14.8 \\
  Wan2.2-5B~\citep{wan2025wan22}        & 5B & 2.739 & 19.4          & 25.0          & 19.4          & 16.3 \\
  Open-Sora-v2~\citep{opensora2}        & 11B & 5.035  & 19.0          & 27.0          & 21.2          & 10.1 \\
  LTX-Video~\citep{HaCohen2024LTXVideo} & 13B & 1.152  & 18.4          & 31.0          & 14.1          & 15.0 \\
  CausalMotion~\citep{zhuang2026causalmotion}  & 13B & 0.395  & 12.7          & 19.0          & 15.0          & 8.1  \\
  Cosmos3-Super~\citep{agarwal2026cosmos} & 64B & - & \textbf{42.7} & - & - & - \\
  MiniMax-H3~\citep{minimax2026h3} & 33B & - & 39.8 & - & - & - \\
  \midrule
  \textbf{Ours (CogVideoX-5B)}         & 20B+5B & 10.02   & \underline{41.9} & \underline{46.7}  & \textbf{53.5} & \textbf{36.0} \\
  \textbf{Ours (Wan2.2-5B)}            & 20B+5B & 4.19 & 39.7 & 45.2 & \underline{53.0} & \underline{35.4} \\
  \bottomrule
  \end{tabular}%
  % }
  \end{center}
  \end{table}

% !TEX root = main.tex

\begin{table}[t]
    \caption{\textbf{PhyGenBench Results.} Comparison of physical-consistency metrics across baselines. \textbf{Bold} and \underline{underlined} denote best and second-best scores.}
    \label{tab:phygenbench_results}
    % \setlength{\tabcolsep}{3.5pt}
    % \resizebox{\linewidth}{!}{%
    \begin{tabular}{lccccccc}
    \toprule
    % \small
    Method & \# param & Mechanics $\uparrow$ & Optics $\uparrow$ & Thermal $\uparrow$ & Material $\uparrow$ & Average $\uparrow$ \\
    \midrule
    VideoCrafter2~\citep{chen2024videocrafter2} & 1.4B & 40.00 & 58.00 & 28.89 & 34.17 & 42.08 \\
    LaVie~\citep{wang2025lavie} & 0.91B & 29.17 & 50.67 & 23.33 & 32.50 & 35.63 \\
    Open-Sora v2~\citep{opensora2} & 11B & 55.00 & 66.67 & 47.78 & 50.83 & 56.25 \\
    LTX-Video~\citep{HaCohen2024LTXVideo} & 13B & 45.83 & 65.33 & 43.33 & 37.50 & 49.37 \\
    DreamWorld~\citep{tan2026dreamworld} & 1.3B & 54.17 & 64.67 & 51.11 & 43.33 & 54.17 \\
    Cosmos3-Nano~\citep{agarwal2026cosmos} & 16B & 57.50 & 75.33 & 48.89 & 46.67 & 58.75 \\
    Cosmos-Predict2.5~\citep{ali2025world} & 2B & 60.00 & 68.00 & 52.22 & 43.33 & 56.87 \\
    PhysVid~\citep{pathak2026physvid} & 1.7B & 44.17 & 60.00 & 38.89 & 33.33 & 45.42 \\
    CausalMotion~\cite{zhuang2026causalmotion} & 13B & \textbf{73.33} & 75.33 & 65.56 & 64.17 & 70.21 \\ 
    % \midrule
    % CogVideoX-2B & 50.83 & 60.00 & 51.11 & 50.83 & 53.75 \\
    % VideoREPA-2B & 50.00 & 63.33 & 50.00 & 44.17 & 52.71 \\
    % LaMo-2B & 47.50 & 56.67 & 48.89 & 45.00 & 50.00 \\
    % Ours-2B & 60.00 & 73.33 & 63.33 & 66.04 & 65.68 \\
    % \midrule
    Wan2.2~\citep{wan2025wan22}& 5B & 50.83 & 65.33 & 41.11 & 40.00 & 50.83 \\
    CogVideoX~\citep{yang2025cogvideox} & 5B & 50.00 & 67.33 & 44.44 & 51.67 & 54.79 \\
    VideoREPA~\citep{zhang2025videorepa} & 5B & 44.17 & 67.33 & 43.33 & 44.17 & 51.25 \\
    LaMo~\citep{jiang2026lamo} & 5B &      45.83 & 65.33 & 46.67 & 44.17 & 51.67 \\
    \midrule
    \textbf{Ours (CogVideoX-5B)} & 20B+5B &     \underline{69.17} & \textbf{80.67} & \underline{81.11} & \textbf{76.67} & \textbf{76.91} \\
    \textbf{Ours (Wan2.2-5B)} & 20B+5B &  67.50 & \underline{79.33} & \textbf{82.22} & \underline{74.17} & \underline{75.62} \\
    \bottomrule
    \end{tabular}%
    % }
\end{table}

\subsection{Experiment Setup}
\label{sec:exp_setup}
We evaluate our proposed system on Physics-IQ Verified~\citep{radsch2026physics,motamed2026physics}, PhyGenBench~\citep{meng2025towards}, PISA Experiments~\citep{li2025pisa}, and our held-out RoboTwin 2.0~\citep{chen2025robotwin} split discussed in Section~\ref{sec:X_exp_dataset}. Physics-IQ Verified evaluates whether generated videos exhibit realistic physical dynamics across phenomena such as gravity, collisions, fluids, materials, lighting, and magnetism. PhyGenBench tests physical commonsense across 27 laws spanning mechanics, optics, thermal phenomena, and material properties, assessing key events, temporal order, and overall naturalness. PISA Experiments focuses on object-dropping scenarios in real and simulated environments, measuring whether generated object trajectories and interactions align with expected physical motion. To evaluate the results of Robotwin2.0 validation against baselines, we apply the EWMScore, a 15-metric adaptation of WorldArena~\citep{shang2026worldarena}, with more details provided in Section~\ref{sec:robotwin_evaluation}.

% \subsection{Baselines}
% \paragraph{SKIP and KeyWorld.} Both papers train a video generator to predict sparse, informative keyframe sequences. Our proposed emphasis is an image-based transition model that predicts the next visual state from the current state, a transition description, and temporal or action conditions. 

% The difference in emphasis would be that KeyWorld and SKIP concentrate expensive rollout generation on informative frames, whereas our work learns an explicit visual transition model that can generate successive future states.

% \paragraph{CausalMotion} explicitly plans both key states and object trajectories. It therefore specifies part of the intermediate motion before video synthesis. Our approach can instead assign more of that continuous dynamics inference to a video model conditioned on the generated states. This is a difference in modeling responsibility, although both approaches use structured intermediate predictions to organize generation.

% \paragraph{CoECT} is particularly close: it already constructs an event chain, progressively edits one event image into the next, and predicts the time spans between events. Our work studies image-based learning of event-conditioned visual state transitions within a key-state-driven world-modeling framework. The distinction would concern the generalization and composition of learned transitions, rather than merely whether a component was fine-tuned.

\subsection{Main Results}

\noindent\textbf{Physics-IQ Verified.}
As shown in Table~\ref{tab:physicsiq-verified}, our method substantially improves physical consistency across both video backbones. It achieves scores of 41.9 with CogVideoX-5B and 39.7 with Wan2.2-5B, surpassing their respective baselines by 11.4 and 20.3 points. Both variants outperform all baselines with reported component metrics on ST-IoU and WS-IoU, with our CogVideoX-based variant reaching 53.5 and 36.0, respectively. Although Wan2.2-14B retains the highest S-IoU, our variants achieve higher overall scores, highlighting improvements beyond spatial agreement alone. Our best variant also surpasses MiniMax-H3 and is only 0.8 points behind Cosmos3-Super. These results demonstrate that augmenting a 5B video backbone with our 20B auxiliary model provides a competitive alternative to increasing the video generator's scale.
% !TEX root = main.tex

\begin{wraptable}{r}{0.55\textwidth}
  \centering
  \caption{\textbf{Pisa-Experiments results}. Results are aggregated over all 421 evaluation videos.}
  \label{tab:pisabench}
  % \setlength{\tabcolsep}{8pt}
  % \resizebox{\linewidth}{!}{%
  \begin{tabular}{lccc}
  \toprule
  Method & L2 $\downarrow$ & CD $\downarrow$ & IoU $\uparrow$ \\
  \midrule
  CausalMotion & 0.1292 & 0.3435 & 0.0966 \\
  CogVideoX-5B & 0.1180 & 0.3017 & 0.1481 \\
  Cosmos-Predict2.5 & 0.1373 & 0.3871 & 0.1467 \\
  Cosmos3-Nano & 0.1250 & 0.3437 & 0.1361 \\
  LaMo-5B & 0.1113 & 0.2888 & \underline{0.1590} \\
  LTX-Video & 0.1181 & 0.3071 & 0.1017 \\
  OpenSoraV2 & 0.1293 & 0.3310 & 0.0846 \\
  \midrule
  \textbf{Ours (CogVideoX-5B)}   & \textbf{0.0949} & \textbf{0.2495} & \textbf{0.1753} \\
  \textbf{Ours (Wan2.2-5B)}      & \underline{0.1010} & \underline{0.2637} & 0.1485 \\
  \bottomrule
  \end{tabular}%
  % }
  \end{wraptable}

\noindent\textbf{PhyGenBench.}
Table~\ref{tab:phygenbench_results} shows that our method improves physical commonsense across all four domains for both video backbones. Our CogVideoX- and Wan-based variants achieve average scores of 76.91 and 75.62, respectively, improving over their corresponding baselines by 22.12 and 24.79 points. They also surpass the strongest baseline CausalMotion by 6.70 and 5.41 points. Our CogVideoX-based variant achieves the highest scores in optics and material properties, while the Wan-based variant leads in thermal phenomena. Although CausalMotion retains the highest mechanics score, these results demonstrate improvements across physical domains and effectiveness across different video backbones.

% !TEX root = main.tex

\begin{figure}[!t]
    \centering
    % \vspace{-2mm}
    \includegraphics[width=\linewidth]{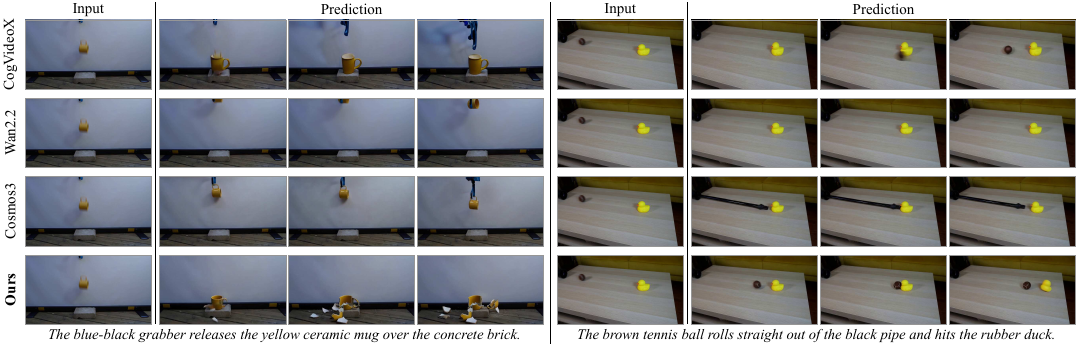}
    \caption{\textbf{Qualitative comparisons on Physics-IQ Verified.} Visualizations of four time-aligned frames baselines and \ours, which depicts mug fracture after impact with a concrete brick (left) and duck displacement following contact with a rolling brown ball (right).}
    \label{fig:physicsiq_qualitative}
    % \vspace{-3mm}
\end{figure}

\noindent\textbf{Pisa-Experiments.}
As shown in Table~\ref{tab:pisabench}, our CogVideoX-based variant achieves the lowest L2 and Chamfer Distance, reducing these errors by 19.6\% and 17.3\%, respectively, relative to CogVideoX-5B-I2V~\citep{yang2025cogvideox}. Our Wan-based variant ranks second on both distance metrics, and both variants outperform all evaluated baselines on L2 and CD. Our CogVideoX-based variant also achieves the highest IoU, improving over its CogVideoX-5B backbone. These results indicate that our approach improves geometric accuracy while maintaining competitive spatial overlap across the 421 evaluation videos.

\begin{figure}[!t]
    \centering
    \includegraphics[width=\linewidth]{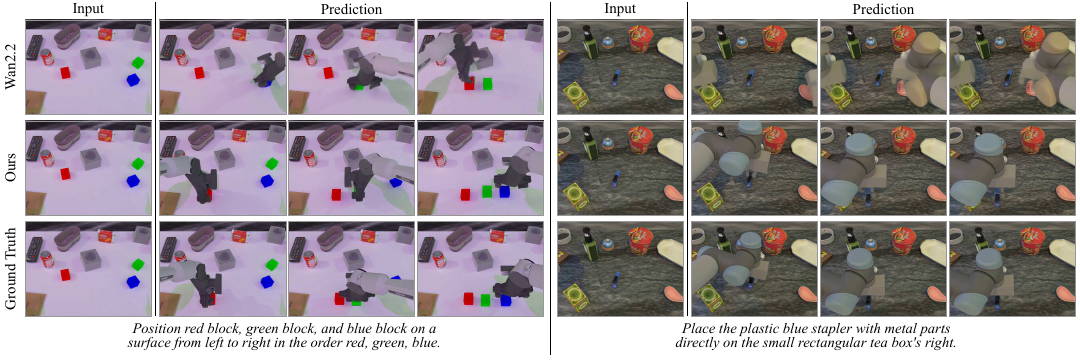}
    \caption{\textbf{Qualitative comparisons on RoboTwin2.0.} Visualizations of Wan2.2~\citep{wan2025wan22} and \ours in comparison against the ground truth.}
    \label{fig:robotwin_qualitative}
\end{figure}

\noindent\textbf{Robotic Manipulation.}
On RoboTwin2 val500, comprising 500 held-out episodes across 50 tasks and five embodiments under an episode/seed-grouped split, both variants outperform all baselines on every displayed metric (Table~\ref{tab:robotwin2_wfm_results} \& Figure~\ref{fig:robotwin_qualitative}). Relative to their respective backbone baselines, our CogVideoX and Wan variants improve EWMScore-15 by 2.47 and 3.37 points, with trajectory-score gains of 43.7\%
and 42.0\%. The Wan variant additionally improves interaction quality and instruction following by 18.2\% and 31.8\%, reaching 62.53 EWMScore-15. These consistent cross-backbone gains support explicit key-state conditioning for improving trajectory fidelity and task-aligned object interactions in generated manipulation videos.

\noindent\textbf{Qualitative Results.}
Figure~\ref{fig:physicsiq_qualitative} compares our method with the baselines on two Physics-IQ Verified examples. In the mug-drop example, our method captures impact followed by fragmentation, whereas the baselines retain an intact mug or leave it suspended above the concrete block. In the ball--duck example, our sequence depicts the ball approaching the duck, making contact, and inducing displacement. The baselines instead exhibit inconsistent ball motion, missing interactions, or spurious changes to the pipe.  By providing explicit intermediate states at planned event times, our method guides video generation toward the intended physical progression. 

% !TEX root = main.tex

% !TEX root = main.tex

\begin{table*}[t]
    \centering                                                                     
      \caption{\textbf{Comparison on the RoboTwin2 val500 benchmark.} EWMScore-15 is the arithmetic mean of all 15 metrics. The best result is  in \textbf{bold}, and the second-best result is \underline{underlined}.}
      \label{tab:robotwin2_wfm_results}
  \setlength{\tabcolsep}{3.5pt}
      \resizebox{\linewidth}{!}{%
      \begin{tabular}{lcccccc}
          \toprule
          \textbf{Method}
          & \textbf{EWMScore} $\uparrow$
          & \textbf{Trajectory} $\uparrow$
          & \textbf{Interaction} $\uparrow$
          & \textbf{Perspectivity} $\uparrow$
          & \textbf{Instruction} $\uparrow$
          & \textbf{Semantic} $\uparrow$ \\
          \midrule
          OpenDW
          & 44.26
          & 0.0220
          & 0.2556
          & 0.6096
          & 0.2080
          & 0.8685 \\

          Ctrl-World
          & 60.16
          & 0.2217
          & 0.5268
          & 0.7776
          & 0.4960
          & 0.8782 \\

          CogVideoX-5B
          & 58.04
          & 0.2301
          & 0.5432
          & 0.7952
          & 0.5180
          & 0.8934 \\

          Wan2.2-5B
          & 59.16
          & 0.2372
          & 0.5088
          & 0.8040
          & 0.4560
          & 0.8866 \\
          \midrule

          \textbf{Ours (CogVideoX-5B)}
          & \underline{60.51}
          & \underline{0.3306}
          & \underline{0.5900}
          & \underline{0.8160}
          & \underline{0.5916}
          & \textbf{0.8975} \\

          \textbf{Ours (Wan2.2-5B)}
          & \textbf{62.53}
          & \textbf{0.3368}
          & \textbf{0.6012}
          & \textbf{0.8448}
          & \textbf{0.6012}
          & \underline{0.8943} \\
          \bottomrule
      \end{tabular}%
      }
  \end{table*}

\subsection{Ablation Studies}

\begin{figure*}[t]
    \centering

    % Measure the image for top alignment with the right-hand column.
    \sbox{\transitionQualBox}{%
        \includegraphics[width=0.6\textwidth]{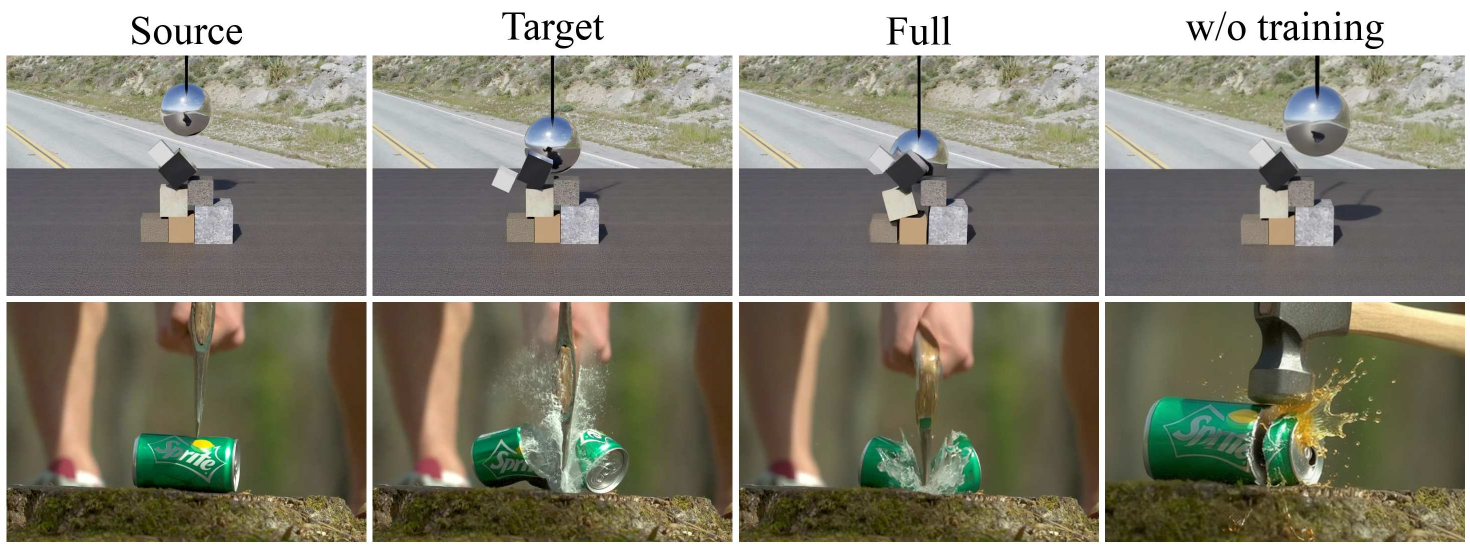}%
    }

    % =====================================================
    % Left: qualitative image and figure caption
    % =====================================================
    \begin{minipage}[t]{0.6\textwidth}
        \vspace{0pt}
        \centering
        \usebox{\transitionQualBox}

        \captionsetup{
            font=small,
            justification=raggedright,
            singlelinecheck=false,
            skip=4pt
        }
        \caption{
            Qualitative comparison between our trained transition model
            and the pretrained Qwen Image Edit.
        }
        \label{fig:keyframe_eval}
    \end{minipage}%
    \hfill
    % =====================================================
    % Right: two compact, non-overlapping tables
    % =====================================================
    \begin{minipage}[t]{0.39\textwidth}
    %     \vspace{12pt}
    %     \centering
    %     \scriptsize
    %     \setlength{\tabcolsep}{2.5pt}
    %     \renewcommand{\arraystretch}{1.12}

        % Inherit caption fonts and label color from the document style.
        \captionsetup{
            % justification=centering,
            singlelinecheck=false,
            position=top,
            skip=2pt
        }

    %     % ---------------- Table 5 ----------------
    %     \captionof{table}{\textbf{Transition state evaluation}. We perform our pipeline on the validation data. Overall we achieve 24.88 in PSNR and 0.863 in SSIM score, outperforming the pretrained Qwen-Image-Edit model}
    %     \label{tab:transition-state-evaluation}

    %     \resizebox{\linewidth}{!}{%
    %     \begin{tabular}{@{}p{0.22\linewidth}
    %         >{\centering\arraybackslash}p{0.14\linewidth}
    %         >{\centering\arraybackslash}p{0.14\linewidth}
    %         >{\centering\arraybackslash}p{0.14\linewidth}
    %         >{\centering\arraybackslash}p{0.14\linewidth}@{}}
    %         \toprule
    %         Model
    %         & PSNR$\uparrow$
    %         & SSIM$\uparrow$
    %         & LPIPS$\downarrow$
    %         & L1$\downarrow$ \\
    %         \midrule
    %         w/o training
    %         & 17.90 & 0.674 & 0.436 & 0.104 \\
    %         \textbf{Ours}
    %         & \textbf{24.88}
    %         & \textbf{0.863}
    %         & \textbf{0.185}
    %         & \textbf{0.038} \\
    %         \bottomrule
    %     \end{tabular}%
    %     }

        \vspace{9pt}

        % ---------------- Table 5 ----------------
        \captionof{table}{\textbf{Ablation on transition model design choices.} Our full model achieves the best scores across different benchmarks}
        \label{tab:ablation_others}
        \centering
        \setlength{\tabcolsep}{1.0pt}
        \small
        % \resizebox{\linewidth}{!}{%
        \begin{tabular}{lcc}
            \toprule
            Variants & Physics-IQ$\uparrow$ & Pisa-IoU$\uparrow$ \\         
            \midrule
            w/o sink & 40.60 & 0.1528 \\            
            w/o time & 41.63 & 0.1617 \\         
            w/o synthetic & 37.91 & 0.1523 \\         
            \midrule
            \textbf{Full model} & \textbf{41.93}  & \textbf{0.1753} \\
            \bottomrule
        \end{tabular}%
        % }
    \end{minipage}

  %   \begin{minipage}[t]{0.40\textwidth}
  %   \vspace{12pt}
  %       \centering
  %       \scriptsize
  %       \setlength{\tabcolsep}{2.5pt}
  %       \renewcommand{\arraystretch}{1.12}

  %       \captionsetup{
  %           font=small,
  %           labelfont=normalfont,
  %           textfont=normalfont,
  %           % justification=centering,
  %           singlelinecheck=false,
  %           position=top,
  %           skip=2pt
  %       }
  %   % \centering
  %   \captionof{table}{Ablation Studies on the variants of the transition model}
  %   \label{tab:physicsiq-ablation}
  %   % \setlength{\tabcolsep}{2.5pt}
  %   \resizebox{\linewidth}{!}{%
  %     \begin{tabular}{lccc}        
  %         \toprule
  %         Variants & Physics-IQ$\uparrow$ & PhyGenBench$\uparrow$ & Pisa$\uparrow$ \\         
  %         \midrule
  %         w/o first-frame sink & 40.60 & &   \\            
  %         w/o time conditioning & 41.63 &  &  \\         
  %         w/o synthetic data & 37.91 & &   \\         
  %         \midrule
  %         \textbf{Full model} & \textbf{41.93} & 76.91 & 0.1523 \\
  %         \bottomrule
  %     \end{tabular}%
  %   }
  % \end{minipage}
  \vspace{9pt}
    \begin{minipage}[t]{\textwidth}
    \centering
    \captionsetup{
            % justification=centering,
            singlelinecheck=false,
            position=top,
            skip=2pt
        }
    \captionof{table}{\textbf{Ablation Studies on learning transition model.} Compared with generating transition states with the pre-trained Qwen Image Edit~\cite{wu2025qwen} model, our learned transition model achieves better performance in both validation sets and multiple benchmarks} % 请根据需要修改 Caption
    \label{tab:ablation_learning}         % 请根据需要修改 Label
    \setlength{\tabcolsep}{7.5pt}
    \resizebox{0.99\linewidth}{!}{%
      \begin{tabular}{lccccccc}
          \toprule
          & \multicolumn{4}{c}{State Transition Evaluation} & \multicolumn{3}{c}{Video Generation Evaluation} \\
      \cmidrule(lr){2-5}
      \cmidrule(lr){6-8}
          Model & PSNR$\uparrow$ & SSIM$\uparrow$ & LPIPS$\downarrow$ & L1$\downarrow$ & Physics-IQ$\uparrow$ & PhyGenBench$\uparrow$ & Pisa-IoU$\uparrow$ \\
          \midrule
          w/o training & 17.90 & 0.674 & 0.436 & 0.104 & 32.60 & 72.50 & 0.1224 \\
          \textbf{Ours} & \textbf{24.88} & \textbf{0.863} & \textbf{0.185} & \textbf{0.038}& \textbf{41.93} & \textbf{76.91} & \textbf{0.1753}  \\
          \bottomrule
      \end{tabular}%
    }
    % \vspace{-8mm}
  \end{minipage}

\end{figure*}

% \input{arxiv/tab/tex/ablation}

% \paragraph{Learned Transition Model.} We first analyze the contribution of transition-model training by comparing the pretrained Qwen-Image-Edit (QIE) model with our trained transition model. We first run our event planning and transition state generation pipeline on our validation set. Figure~\ref{fig:keyframe_eval} and Table~\ref{tab:ablation_learning} demonstrate that our pipeline achieves effective construction of the transition states: PSNR increases from 17.90 to 24.88 (+6.98 dB), while SSIM improves from 0.674 to 0.863. Meanwhile, LPIPS decreases from 0.436 to 0.185 and L1 decreases from 0.104 to 0.038, corresponding to relative reductions of 57.6\% and 63.5\%, respectively. These gains translate into stronger physical video generation: Physics-IQ improves from 32.60 to 41.93 (+9.33 points), PhyGenBench from 72.50 to 76.91 (+4.41 points), and Pisa-IoU from 0.1224 to 0.1753 (+0.0529). This result demonstrates that learning explicit visual state transitions is essential for producing physically consistent downstream videos.
\paragraph{Learned Transition Model.}
  We compare pretrained Qwen-Image-Edit (QIE) with our trained transition model through the event-planning and transition-state
  generation pipeline (Figure~\ref{fig:keyframe_eval}; Table~\ref{tab:ablation_learning}). Metrics are computed against
  reference states, excluding the initial input image. On validation states, PSNR increases from 17.90 to 24.88 and SSIM from
  0.674 to 0.863, while LPIPS and L1 decrease by 57.6\% and 63.5\%, respectively. Both perceptual similarity and pixel-level
  accuracy improve. These gains accompany improvements in downstream video generation: Physics-IQ rises from 32.60 to 41.93,
  PhyGenBench from 72.50 to 76.91, and Pisa-IoU from 0.1224 to 0.1753. Together, these results support learning task-specific
  transitions and using their predicted states to condition dense video synthesis.

  \paragraph{Design Choice Analysis.}
  Table~\ref{tab:ablation_others} examines first-frame anchoring, temporal conditioning, and synthetic data. Removing the
  first-frame anchor lowers Physics-IQ/Pisa-IoU to 40.60/0.1528, supporting persistent conditioning on the initial visual
  reference during recursive prediction. Without time conditioning, Physics-IQ decreases modestly to 41.63, whereas Pisa-IoU
  falls to 0.1617, indicating benchmark-dependent benefits. Excluding synthetic data yields 37.91/0.1523 and the largest
  Physics-IQ drop of 4.02 points. These comparisons favor retaining all three design choices, although differences in training
  mixtures and budgets limit strict causal attribution to individual components.

% !TEX root = main.tex

\section{Conclusion and Limitation}
We presented a physical world modeling framework that explicitly learns visual state transitions and uses their timed
  predictions to condition dense video generation. VLM-assisted annotation constructs event-aligned supervision, while a
  pretrained event planner provides transition specifications for open-domain inference. Across physical-video and robotic
  manipulation benchmarks, the resulting intermediate-state representation improves physical-consistency and manipulation-
  video metrics over corresponding video backbones. However, the framework remains a learned visual predictor rather than an
  explicit physics simulator: plausible states do not guarantee correct trajectories, contact dynamics, or conservation laws.
  Since training uses observed states but inference conditions on recursively predicted states, errors may accumulate over
  long horizons. The pipeline also introduces additional planning and image-generation costs and depends on the quality of
  model-assisted annotations.

\bibliographystyle{ACM-Reference-Format}
\bibliography{iclr2027_conference}

\clearpage
\appendix
% !TEX root = main.tex

% !TEX root = main.tex

\section{Additional Training Details}

\subsection{Datasets}
\label{sec:X_exp_dataset}

We use complementary datasets for learning visual state transitions, training conditional video dynamics, and studying robotic manipulation. Table~\ref{tab:dataset_summary} summarizes the prepared corpora. The reported sizes refer to
curated data before reserving internal validation samples, rather than to the nominal sizes of the upstream datasets.

\paragraph{WISA.}
WISA~\citep{wang2025wisa} provides real-world videos covering diverse physical phenomena. Our transition corpus retains 59,298 episodes across 17 annotated categories, including collisions, elastic motion, deformation, liquid and gas
motion, phase changes, combustion, reflection, and refraction. WISA contributes 160,672 adjacent-state pairs and is also a source for the separately curated video-dynamics corpus described below.

\paragraph{NVIDIA PhysicalAI.}
We use 39,873 episodes from NVIDIA PhysicalAI's synthetic physical-interaction scenes~\citep{nvidia_physicalai-worldmodel-synthetic-physical-interaction-scenes_2026}. Our selected subset covers six scenario families: billiards, bowling,
dominoes, falling objects, objects rolling on ramps, and wrecking-ball interactions. These simulated videos provide examples of contact, momentum transfer, falling, and chained rigid-body interactions that complement the
visual diversity of WISA. Annotation produces 116,358 adjacent-state pairs from this source; we use the selected subset, not the entire upstream PhysicalAI collection.

\paragraph{PhyCo/Kubric.}
We additionally curate 31,109 episodes from PhyCo's Kubric-based simulations~\citep{narayanan2026phyco}. The prepared subset comprises four scenario families: rigid ball drops, soft-body ball drops, ball--wall collisions, and deformable-cube interactions. It contributes 76,810 adjacent-state pairs, providing supervision for changes around impact, rebound, and deformation. Together, WISA, PhysicalAI, and PhyCo/Kubric supply 130,280 video episodes and 353,840 event-aligned transition pairs.

\paragraph{PICA-100K.}
PICA-100K~\citep{pu2026picabench} supplies auxiliary instruction-based image-editing supervision. It consists of synthetic, video-derived editing examples rather than trajectories from a physics simulator. Our prepared manifest contains 105,085 source--target image pairs, each paired with an editing instruction. These examples extend supervision to instruction-driven
changes without requiring a temporal annotation. We therefore use them as text-conditioned image pairs with the time condition omitted, rather than assigning an artificial elapsed time or counting them as additional video episodes.

\paragraph{OpenVid-1M and conditional video dynamics.}
OpenVid-1M~\citep{nan2025openvid} provides captioned, open-domain videos that complement WISA for training the conditional dynamics model. We annotate sparse states and retain WISA/OpenVid sequences for which all selected frames
pass the state-description consistency check described in Section~\ref{sec:annotation_planning}. This additional filtering is specific to this video-training path and is not shared by all transition-pair sources. The prepared CogVideoX~\citep{yang2025cogvideox} corpus contains 69,344 episodes (35,114 WISA and 34,230 OpenVid), represented as 49-frame clips at 720$\times$480. The Wan2.2~\citep{wan2025wan22} corpus contains same number of episodes, represented as 81-frame clips at 832$\times$480. Each backbone reserves 2,000 episodes for internal validation.
Training uses observed state images with their temporal locations as conditions and the corresponding full video as the target.

\paragraph{RoboTwin2.0.}
For robotic manipulation, we use 124,650 complete simulated episodes from RoboTwin2.0~\citep{chen2025robotwin}, spanning 50 tasks and five robot embodiments: ALOHA-AgileX, ARX-X5, Franka, UR5, and Piper. This snapshot contains 11,150 clean and 113,500 domain-randomized episodes. We construct a deterministic 90/10 split into 112,185 training and 12,465 validation episodes, keeping episodes with the same task and simulation seed in the same split across embodiments and clean/randomized configurations. This prevents the same task--seed group from appearing on both sides. For the reported \emph{val500} evaluation, we select 500 held-out episodes covering all 50 tasks and all five embodiments; every selected episode is domain-randomized. Thus, the evaluation measures generalization to unseen episodes of known tasks, not to held-out task categories.

% !TEX root = main.tex

\begin{table}[t]
    \centering
    \caption{Prepared datasets used in our experiments.}
    \label{tab:dataset_summary}
    \small
    \setlength{\tabcolsep}{5pt}
    \renewcommand{\arraystretch}{1.12}
    \begin{tabular*}{\linewidth}{@{\extracolsep{\fill}}llr@{}}
        \toprule
        \textbf{Dataset} & \textbf{Type} & \textbf{Size} \\
        \midrule
        \multicolumn{3}{@{}l}{\textit{Transition-model supervision}} \\
        WISA-80K          & Real       & 59,298 episodes \\
        NVIDIA PhysicalAI & Simulation & 39,873 episodes \\
        PhyCo/Kubric     & Simulation & 31,109 episodes \\
        PICA-100K        & Synthetic & 105,085 image pairs \\
        \midrule
        \multicolumn{3}{@{}l}{\textit{Conditional video-dynamics supervision}} \\
        WISA-80K         & Real       & 35,114 episodes \\
        OpenVid-1M       & Real       & 34,230 episodes \\
        \midrule
        \multicolumn{3}{@{}l}{\textit{Robotic manipulation}} \\
        RoboTwin2.0      & Simulation & 112,185 train; 12,465 val \\
        RoboTwin2.0 val500 & Simulation & 500 validation episodes \\
        \bottomrule
    \end{tabular*}
\end{table}

\subsection{Event Planning and Data Annotation Pipeline}
We construct transition state frame annotations through a two-stage vision--language model (VLM) annotation pipeline, followed by VQA-based verification. For each video, candidate frames are arranged into a temporally ordered storyboard and provided alongside the video caption, available source metadata, and a candidate table specifying frame identifiers and timestamps. 

\paragraph{Stage I: Event Planning.} In (Table~\ref{tab:keyframe-event-planning-prompt}), the annotator identifies the main objects and physical or visual process, characterizes the event structure, and decomposes the observed progression into a minimal sequence of visible subevents. It then constructs a state plan grounded in candidate frames, prioritizing physical state changes while also allowing camera motion that reveals new content or distinct viewpoints. Clips without useful visible state or view progression are flagged as unsuitable for event-based annotation. 

\paragraph{Stage II: Transition State Frame Selection and Annotation.} In (Table~\ref{tab:keyframe-selection-prompt}), the annotator uses this event plan to select a compact, chronologically ordered set of event-defining transition state frames, rather than generic beginning--middle--end coverage. Selection prioritizes the earliest frame in which a meaningful state or relation is clearly established, while avoiding near-duplicates and redundant intermediate frames. Each selected transition state frame receives a state description, a selection rationale, and a confidence score; adjacent  transition state frames additionally receive a textual description of the intervening transition. 

\paragraph{Stage III: Transition State Frame State Verification.} In (Table~\ref{tab:qwen-vqa-keyframe-judge-prompt}) uses Qwen3-VL~\citep{bai2025qwen3} as a separate VQA judge to assess whether each annotated state is supported by the visual evidence. The first transition state frame is evaluated on its own, whereas subsequent transition state frames are evaluated together with the immediately preceding transition state frame as a reference for changes or motion. Conditioned on the video caption, physical category, and claimed state, the judge returns a binary match verdict, a confidence score, and a short justification, rejecting claims that are not clearly visible. These verdicts are aggregated at the video level to support downstream annotation-quality filtering.

% \paragraph{Transition annotation and sampling.}
% The offline pipeline in Section~\ref{sec:annotation_planning} selects
% chronologically ordered states from each source video and describes the
% changes between adjacent states. Target images are extracted from the videos,
% not synthesized by the annotator, and relative temporal offsets are derived
% from their frame indices. Transition-model images are prepared at
% 720$\times$480, and pairs from the same video retain a shared identifier for
% internal train--validation splitting. We sample three supervision streams
% with equal probability: annotated state pairs with both transition text and
% time; pairs formed by an annotated source frame and a randomly sampled future
% frame, with time but no transition text; and PICA editing pairs with text but
% no time. The random-future-frame stream samples valid offsets of 5--25
% processed frames. This mixture combines event-aligned supervision with
% additional temporal coverage and instruction-based editing.

\subsection{Training Configurations}
\label{sec:X_training_config}
For transition model, we fine-tune Qwen-Image-Edit~\citep{wu2025qwen} as a time-conditioned Markov model using paired examples from the comprehensive multi-source training corpus. Each example consists of the previous ground-truth state image, the next state image, and a text instruction describing their transition. Images are encoded with a frozen VAE, while the frozen vision-language encoder processes the instruction and conditioning image. The transformer receives concatenated noisy target latents and clean source latents. Relative transition state frame timing is represented through temporal rotary positional embeddings, assigning positions zero and ($\Delta t$) to source and target tokens, respectively. We optimize a flow-matching MSE objective on target tokens only, keeping the source image uncorrupted. The complete transformer is fine-tuned with AdamW~\citep{loshchilov2017decoupled} at a learning rate of $10^{-5}$, using 100 warm-up steps. During auto-regressive inference, each generated transition state images becomes the semantic condition for the next transition. For conditional dynamic model training, we fine-tune CogVideoX-5B-I2V~\citep{yang2025cogvideox} and Wan2.2-TI2V-5B~\citep{wan2025wan22} using AdamW~\citep{loshchilov2017decoupled} with learning rate equals to $10^{-5}$ and weight decay 0.01, keeping the VAE and text encoder frozen. Transition states latents are patchified using the pretrained projection, and temporally aligned with video tokens. The tokens are prepended with video stream for joint text–frame–video attention. Only video outputs are supervised, using velocity prediction for CogVideoX and flow matching for Wan. The transition model is trained with 64 A100 GPUs for 5 epochs with per-GPU batch size of 1 and the conditional dynamics model is trained with 32 GPUs for 5000 steps with per-GPU batch size of 2.

% !TEX root = main.tex

\begin{table}[!t]
\caption{\textbf{Event planning prompt.}
Prompt used to identify the visible physical process, ordered subevents,
and event-relevant states before transition state selection.}
\label{tab:keyframe-event-planning-prompt}

\centering
% Horizontal padding between the text and the left/right borders.
\setlength{\tabcolsep}{12pt}

% Subtract the padding and border widths so the table remains within \textwidth.
\begin{tabular}{|p{\dimexpr\textwidth-2\tabcolsep-2\arrayrulewidth\relax}|}
\hline

\begin{minipage}[t]{\linewidth}
\vspace{0.8em}

{\ttfamily\footnotesize
\raggedright

You are labeling transition state frames for training a physics-aware video generation model.

\medskip

You will receive \{N\} storyboard images, one per sample. Each storyboard
image contains labeled candidate frames.

\medskip

For each sample, do this in order:

\begin{enumerate}[
    leftmargin=2.1em,
    itemsep=0.45em,
    topsep=0.4em,
    parsep=0pt,
    label=\arabic*.
]

    \item Determine whether the video contains a visible physical state change.

    \item Identify the main objects, scene elements, and main physical or
    visual process.

    \item Infer event\_structure using a concise general label, such as
    single\_transition, sequential\_transitions, repeated\_motion,
    camera\_motion, scene\_change, or unclear.

    \item Write ordered\_subevents as the minimal ordered list of visible
    subevents. Merge simultaneous changes into one subevent and split clearly
    sequential changes into separate subevents.

    \item Build state\_plan around moments that matter for transition state selection.
    State descriptions should mention visible states or relations such as
    onset of motion, contact or collision, deformation, liquid or granular
    flow, opening or closing, placement, stacking, release, stable result,
    camera reveal, or view change.

    \item Reject clips that are mostly static and have no useful visible
    state/view progression.

\end{enumerate}

\medskip

Return strict JSON only with one item per sample\_id.

}

\vspace{0.8em}
\end{minipage}

\\
\hline
\end{tabular}
\end{table}

\subsection{Training and Inference Protocol}
\label{sec:X_training_protocol}
The transition and dynamics models are trained separately. For dynamics training, conditioning states are extracted from the training video and paired with their temporal indices; the dense video supplies the prediction target. At inference, these observed state conditions are replaced by the transition model's generated states. Thus, the training and inference interfaces agree in their structure, while the quality of the state conditions can differ.

Inference consists of state event planning, successive state generation, and dense video sampling conditioned on the resulting timed memories. The original task context is retained by the video generator. Generated videos are subsequently formatted for benchmark evaluation. The benchmark evaluator is external to the generative model and supplies no gradients to either learned component.

% !TEX root = main.tex

\begin{table}[t]
    \centering
      \caption{\textbf{Results on VideoPhy.} Model size, full-benchmark generation cost (EFLOPs), and semantic adherence and physical common sense metrics are compared across multiple baseline methods, with \textbf{bold} and \underline{underlined} values indicating the best and second-best scores, respectively. Our results are reported with text-to-video (T2V) setting.}
      \label{tab:X_videophy_results}
      % \setlength{\tabcolsep}{3.5pt}
      % \small
      \resizebox{\linewidth}{!}{%
      \begin{tabular}{lcccc cc cc cc}
          \toprule
          & & 
          & \multicolumn{2}{c}{Solid--Solid}
          & \multicolumn{2}{c}{Solid--Fluid}
          & \multicolumn{2}{c}{Fluid--Fluid} 
          & \multicolumn{2}{c}{Overall} \\
          \cmidrule(lr){4-5}
          \cmidrule(lr){6-7}
          \cmidrule(lr){8-9}
          \cmidrule(lr){10-11}
          Method & \# params & EFLOPs$\downarrow$ & SA $\uparrow$ & PC $\uparrow$ & SA $\uparrow$ & PC $\uparrow$ & SA $\uparrow$ & PC $\uparrow$ & SA $\uparrow$ & PC $\uparrow$ \\
          \midrule
          % \multicolumn{9}{l}{\textit{General video-generation baselines}} \\
          VideoCrafter2     & 1.4B   & 0.441  & 27.3          & 25.2          & 47.3          & 24.7          & 49.1          & 30.9          & 39.2          & 25.9 \\
          Open-Sora v2      & 11B    & 9.471  & 53.9          & 13.3          & 72.6          & 24.0          & 65.5          & 27.3          & 63.7          & 20.2 \\
          LTX-Video         & 13B    & 1.379  & 39.9          & 17.5          & 70.6          & 22.6          & 49.1          & 25.5          & 54.4          & 20.9 \\
          DreamWorld        & 1.3B   & 9.830  & 58.7          & 12.6          & 74.0          & 27.4          & 70.9          & 32.7          & 67.2          & 22.1 \\
          Cosmos3-Nano      & 16B    & 14.033 & \textbf{67.1} & 13.3          & \textbf{82.2} & 23.3          & \underline{78.2}          & 25.5          & \textbf{75.3} & 19.5 \\
          Cosmos-Predict2.5 & 2B     & 4.448  & 55.2          & 20.3          & 74.7          & 26.0          & 74.6          & 40.0          & 66.6          & 25.9 \\
          WISA              & 14B    & 59.272 & 58.0          & 14.7          & \underline{78.8}          & 18.5          & \textbf{80.0} & 25.5          & \underline{70.5}          & 18.0 \\
          PhysVid           & 1.7B   & 11.090 & 45.5          & \textbf{21.0} & 74.7          & 30.1          & 76.4          & 40.0          & 62.8          & 27.9 \\
          Wan2.2            & 5B     & 1.753  & 52.5          & 14.7          & 76.0          & 20.6          & 61.8          & 25.5          & 64.0          & 18.9 \\
          CausalMotion      & 13B    & 0.502  & 49.7          & 9.8           & 65.8          & 17.1          & 63.6          & 20.0          & 58.7          & 14.5 \\
          CoECT         & 20B+5B     &   -    & \underline{61.5}        & 9.8           & 80.1          & 17.8          & 69.1          & 27.3          & 70.6          & 16.0 \\
          CogVideoX         & 2B     & 4.236  & 51.8          & 14.7          & 75.3          & 30.1          & 70.9          & 54.6          & 64.8          & 27.6 \\
          VideoREPA         & 2B     & 4.236  & 51.8          & 12.6          & 76.0          & 29.5          & 67.3          & 56.4          & 64.5          & 26.7 \\
          LaMo              & 2B     & 4.236  & 58.7          & 16.8          & 74.7          & \textbf{32.2}          & 69.1          & \underline{67.3}          & 67.2          & \textbf{31.4} \\
          \toprule
          % \hline
          \textbf{Ours (CogVideoX-2B-T2V)}              & 20B+2B & 10.170 & 59.4          & \underline{19.6} & 78.1          & \textbf{32.2}          & \underline{78.2}          & \textbf{69.1} & 70.4          & \textbf{32.9} \\
          % \midrule
          % CogVideoX         & 5B     & 11.536 & 65.7          & 18.9          & 76.0          & 32.9          & 70.9          & 61.8          & 70.9          & 31.7 \\
          % VideoREPA         & 5B     & 11.536 & 57.3          & 28.0          & 80.1          & 39.0          & 78.2          & 69.1          & 70.4          & 39.2 \\
          % LaMo              & 5B     & 11.536 & 62.9          & 26.6          & 80.8          & 41.1          & 78.2          & \textbf{78.2} & 73.0          & 41.0 \\
          % Ours              & 20B+5B & 22.850 & 64.3          & 18.2          & 81.5          & 34.9          & 76.4          & 69.1          & 73.6          & 33.4 \\
          \bottomrule
      \end{tabular}%
      }
      
  \end{table}

\subsection{Robotic Manipulation Evaluation}
\label{sec:robotwin_evaluation}

We evaluate robotic manipulation videos using a 15-metric adaptation of WorldArena~\citep{shang2026worldarena}, denoted EWMScore-15. The retained metrics cover visual quality (image quality, aesthetic quality, and JEPA similarity), motion quality (dynamic degree, flow score, and motion smoothness), content consistency (subject, background, and photometric consistency), physics adherence (interaction quality and trajectory accuracy), 3D accuracy (depth accuracy and perspectivity), and controllability (instruction following and semantic alignment). We exclude Action Following, which requires generating videos under multiple alternative instructions from the same initial frame and is outside our single-instruction evaluation protocol. Using the evaluator's metric-specific normalization and aggregation rules, we obtain dataset-level scores $s_j \in [0,1]$, with higher values indicating better performance, and compute $\mathrm{EWMScore\text{-}15} = \frac{100}{15}\sum_{j=1}^{15}s_j$. All retained metrics receive equal weight. JEPA similarity is evaluated over the complete generated and reference video collections rather than averaged over individual episodes.

% !TEX root = main.tex

\begin{table}[!t]
\caption{\textbf{Transition state frame selection prompt.}
Prompt used to select event-relevant transition state frame from candidate frames.}
\label{tab:keyframe-selection-prompt}

\centering
\setlength{\tabcolsep}{12pt}

\begin{tabular}{|p{0.97\textwidth}|}
\hline

\begin{minipage}[t]{0.94\textwidth}
\vspace{0.8em}

{\ttfamily\footnotesize
\raggedright

Select transition state frames from the candidate frames for each sample.

\medskip

Decision procedure:

\begin{itemize}[
    leftmargin=1.7em,
    itemsep=0.35em,
    topsep=0.4em,
    parsep=0pt
]

    \item First read the caption, source metadata,
    event\_plan.event\_structure,
    event\_plan.ordered\_subevents, and
    event\_plan.state\_plan.

    \item Choose transition state frames for distinct visible states in the main event,
    not for generic beginning/middle/end coverage.

    \item For a single clear transition, choose the earliest candidate where
    the defining changed state is visible; include an earlier setup state
    only when it is needed to understand the change.

    \item For sequential transitions, select one frame per visible subevent
    at the earliest candidate where that subevent's state is clearly
    established.

    \item For contact, collision, attachment, containment, placement,
    stacking, opening, or closing, choose the first candidate where the
    relation is visibly established, not a later redundant proof frame.

    \item For deformation, breakage, spilling, pouring, smoke, fire, liquid,
    granular motion, or other continuous processes, cover the onset and
    the most informative result or peak state. Add an intermediate keyframe
    only when the process has a materially different middle state.

    \item For object motion, choose frames where the object's state or relation
    has changed meaningfully. Avoid frames that differ only by a small
    position shift unless that shift completes the event.

    \item For camera motion without object state change, choose frames that
    capture distinct viewpoints or newly revealed content.

\end{itemize}

\medskip

\textbf{Rules:}

\begin{itemize}[
    leftmargin=1.7em,
    itemsep=0.35em,
    topsep=0.4em,
    parsep=0pt
]

    \item Choose only candidate\_id values in the sample's candidate table.

    \item Do not select static initial frames, near-duplicates, or redundant
    middle frames unless they represent a distinct visible state.

    \item A transition state frame should be an event-defining visual state:
    onset, contact, relation established, state transformed, view changed,
    peak action, or stable result.

    \item Selected frames must be chronological.

    \item Each selected frame must represent a distinct event-relevant
    visible state or relation.

    \item For each transition state frame provide:
    slot, candidate\_id, frame\_index, time\_sec, state\_text,
    selection\_reason, confidence.

    \item Also provide transition\_text between adjacent transition state frames.

    \item In coverage\_notes, explicitly write:

    event\_structure=\textless label\textgreater;

    selected\_subevents=
    \textless comma-separated selected visible states\textgreater;

    omitted=
    \textless why any expected state was not selected or `none'\textgreater.

\end{itemize}

}

\vspace{0.8em}
\end{minipage}

\\
\hline
\end{tabular}
\end{table}

% !TEX root = main.tex

\begin{table}[!t]
\caption{\textbf{Transition state frame verification prompt.}
Prompt used by Qwen3-VL to judge whether an extracted transition frame state visibly
matches its annotated physical state.}
\label{tab:qwen-vqa-keyframe-judge-prompt}

\centering
% Horizontal padding between the text and the left/right borders.
\setlength{\tabcolsep}{18pt}

% Subtract the padding and border widths so the table remains within \textwidth.
\begin{tabular}{|p{\dimexpr\textwidth-2\tabcolsep-2\arrayrulewidth\relax}|}
\hline

\begin{minipage}[t]{\linewidth}
\vspace{0.8em}

{\ttfamily\footnotesize
\raggedright

\textbf{System prompt}

\medskip

You are a strict quality inspector for a physics-focused video dataset.
Judge only what is actually visible in the given frame(s). Be skeptical:
if the described state is not clearly visible, answer false. Always reply
with a single JSON object and nothing else.

\medskip

\textbf{User prompt for the first transition state frame}

\medskip

This image is transition state frame \{slot\_pos\} of \{total\} extracted from a video.

Video context: physical category = ``\{label\}''; caption = ``\{caption\}''

The transition state frame annotation claims the scene has reached this state:
``\{state\_text\}''

Question: does the image show that the scene has actually reached the
claimed state? Judge the physical state of the scene, not image style or
camera work.

Reply with JSON only:
\{\{``match'': true or false, ``confidence'': 0.0--1.0,
``reason'': ``\textless at most 20 words\textgreater''\}\}

\medskip

\textbf{User prompt for a subsequent transition state frame}

\medskip

You are shown two transition state frames from the same video. Image 1 is the PREVIOUS
transition state frame (reference only). Image 2 is the CURRENT transition state frame
\{slot\_pos\} of \{total\}.

Video context: physical category = ``\{label\}''; caption = ``\{caption\}''

The annotation claims that by the CURRENT transition state frame the scene has reached
this state:
``\{state\_text\}''

Question: comparing against image 1 where the claim describes change or
motion, does image 2 show that the scene has actually reached the claimed
state? Judge the physical state of the scene, not image style or camera
work.

Reply with JSON only:
\{\{``match'': true or false, ``confidence'': 0.0--1.0,
``reason'': ``\textless at most 20 words\textgreater''\}\}

}

\vspace{0.8em}
\end{minipage}

\\
\hline
\end{tabular}
\end{table}

% \input{arxiv/tab/tex/X_inference_event_planner}

% !TEX root = main.tex

\section{Additional Results on Benchmarks}
\begin{figure}[!t]
  \centering
  \includegraphics[width=\linewidth]{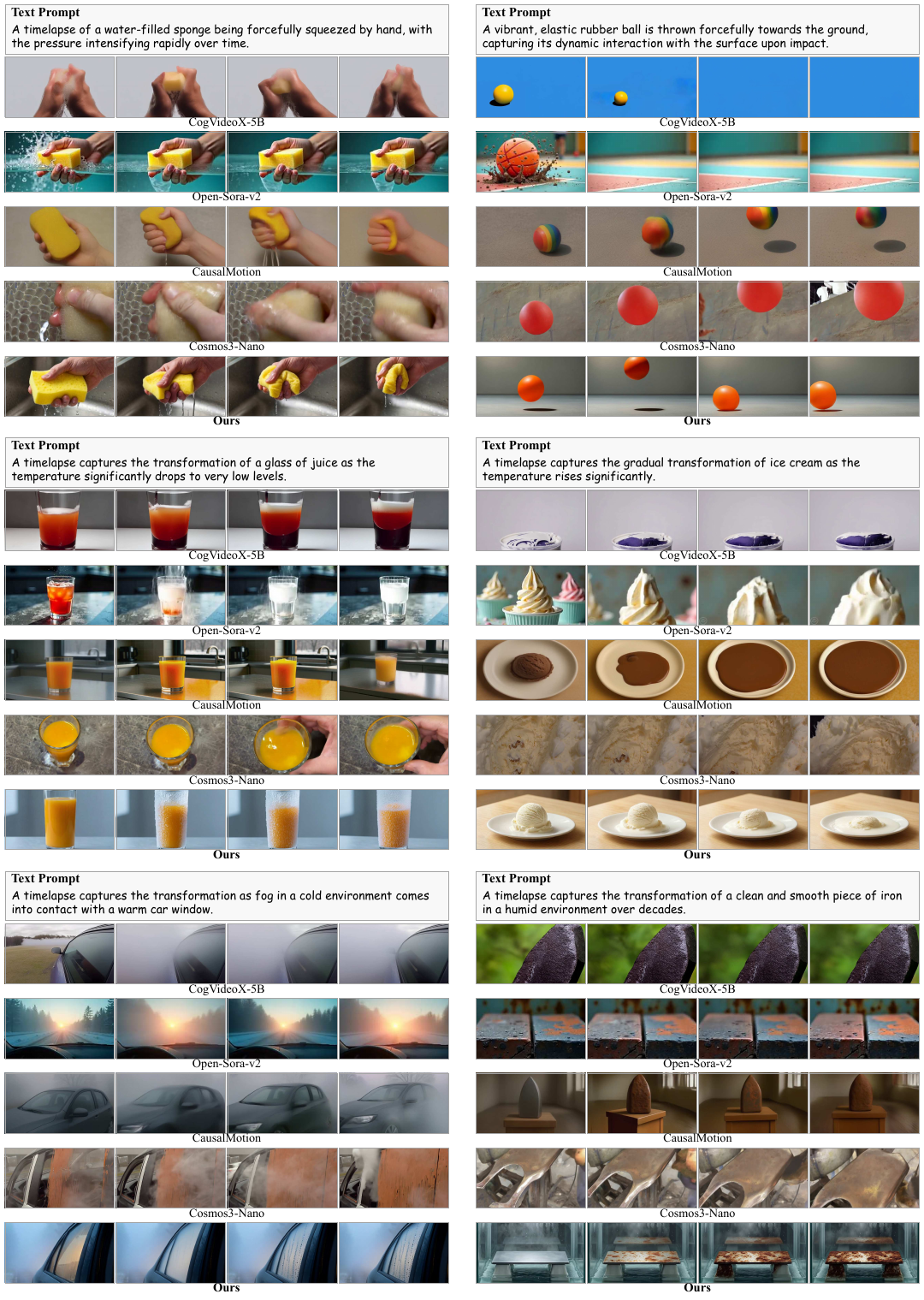}
  \caption{\textbf{Qualitative comparisons on PhyGenBench}. Each block shows the text prompt and four frames from CogVideoX-5B, Open-Sora-v2, CausalMotion, Cosmos3-Nano, and our method.}
  \label{fig:X_phygenbench_qualitative}
\end{figure}

Table~\ref{tab:X_phygenbench_results} and table~\ref{tab:X_videophy_results} demonstrate additional quantitative results on PhyGenBench~\citep{meng2025towards} and VideoPhy~\citep{bansal2025videophy}, another benchmark that evaluates physical common-sence. Compared with other baseline methods, our method uses a substantially larger transition model (20B) together with only a lightweight 2B video backbone, yet remains competitive with, and often surpasses, baselines using 11--16B video models. Since the 2B version of CogVideoX~\citep{yang2024cogvideox} doesn't have an image-to-video (I2V) branch, our results are reported with text-to-video (T2V) setting. On PhyGenBench~\citep{meng2025towards}, it achieves the second-best average score (65.68) and the best Material score (66.04). On VideoPhy~\citep{bansal2025videophy}, it obtains the best overall physical common-sense precision (32.9), the best Fluid--Fluid precision (69.1), and competitive overall semantic accuracy (70.4), while using a lightweight 2B video generator. This suggests that allocating capacity to explicit transition-state prediction can be more effective than scaling the video backbone alone: the transition model supplies structured, physically meaningful intermediate states, allowing a compact dynamics model to render temporally coherent videos at substantially lower model complexity. In addition, Figure~\ref{fig:X_phygenbench_qualitative} visualizes qualitative comparisons with our baseline methods.
% !TEX root = main.tex

\begin{table}[t]
    \caption{\textbf{Results on PhyGenBench.} Model size, and different physical-consistency metrics are compared across multiple baseline methods, with \textbf{bold} and \underline{underlined} values indicating the best and second-best scores, respectively. Our results are reported with text-to-video (T2V) setting.}
    \label{tab:X_phygenbench_results}
    \resizebox{\linewidth}{!}{%
    \begin{tabular}{lccccccc}
    \toprule
    Method & \# param & Mechanics $\uparrow$ & Optics $\uparrow$ & Thermal $\uparrow$ & Material $\uparrow$ & Average $\uparrow$ \\
    \midrule
    Open-Sora v2~\citep{opensora2} & 11B & 55.00 & 66.67 & 47.78 & 50.83 & 56.25 \\
    LTX-Video~\citep{HaCohen2024LTXVideo} & 13B & 45.83 & 65.33 & 43.33 & 37.50 & 49.37 \\
    Cosmos3-Nano~\citep{agarwal2026cosmos} & 16B & 57.50 & 75.33 & 48.89 & 46.67 & 58.75 \\
    CausalMotion~\cite{zhuang2026causalmotion} & 13B & \textbf{73.33} & \textbf{75.33} & \textbf{65.56} & \underline{64.17} & \textbf{70.21} \\ 
    % \midrule
    % CogVideoX-2B & 50.83 & 60.00 & 51.11 & 50.83 & 53.75 \\
    % VideoREPA-2B & 50.00 & 63.33 & 50.00 & 44.17 & 52.71 \\
    % LaMo-2B & 47.50 & 56.67 & 48.89 & 45.00 & 50.00 \\
    \toprule
    \textbf{Ours (CogVideoX-2B-T2V)} & 20B+2B & \underline{60.00} & \underline{73.33} & \underline{63.33} & \textbf{66.04} & \underline{65.68} \\
    % \midrule
    % Wan2.2~\citep{wan2025wan22}& 5B & 50.83 & 65.33 & 41.11 & 40.00 & 50.83 \\
    % CogVideoX~\citep{yang2025cogvideox} & 5B & 50.00 & 67.33 & 44.44 & 51.67 & 54.79 \\
    % VideoREPA~\citep{zhang2025videorepa} & 5B & 44.17 & 67.33 & 43.33 & 44.17 & 51.25 \\
    % LaMo~\citep{jiang2026lamo} & 5B &      45.83 & 65.33 & 46.67 & 44.17 & 51.67 \\
    % \textbf{Ours (CogVideoX)} & 20B+5B &     \underline{69.17} & \textbf{80.67} & \underline{81.11} & \textbf{76.67} & \textbf{76.91} \\
    % \textbf{Ours (Wan)} & 20B+5B &  67.50 & \underline{79.33} & \textbf{82.22} & \textbf{74.17} & \underline{75.62} \\
    \bottomrule
    \end{tabular}%
    }
\end{table}

% !TEX root = main.tex

\end{document}